\documentclass[journal]{IEEEtran}

\usepackage{pdfpages}
\usepackage{colortbl}
\usepackage{enumitem}
\usepackage{float} 
\usepackage{tabularx}
\usepackage{multirow, makecell}
\usepackage{booktabs}
\usepackage{glossaries}
 \usepackage[linesnumbered, ruled,lined]{algorithm2e}
\usepackage{geometry}
\usepackage{changepage}
\usepackage{graphicx}
\usepackage{amsmath}
\usepackage{amssymb}
\usepackage{algorithm2e}
\usepackage{arydshln}
\usepackage{tabu}
\usepackage{array}
\usepackage{arydshln}
\usepackage{caption}
\usepackage{romannum}
\usepackage{dsfont}

\newcolumntype{x}[1]{>{\centering\arraybackslash\hspace{0pt}}p{#1}}
\newcolumntype{y}[1]{>{\arraybackslash\hspace{0pt}}p{#1}}
\newcolumntype{M}[1]{>{\centering\arraybackslash}m{#1}}
\newcolumntype{R}[1]{>{\raggedleft\let\newline\\\arraybackslash\hspace{0pt}}m{#1}}
\newcolumntype{C}[1]{>{\centering\arraybackslash}X{#1}}
\newcolumntype{Y}{>{\centering\arraybackslash}X}

\DeclareCaptionLabelFormat{custom}{\textbf{#2}}

\let\oldthefigure\thefigure
\let\oldtheequation\theequation
\let\oldthetable\thetable

\renewcommand\thefigure{Fig. \oldthefigure}
\renewcommand\theequation{Eq. \oldtheequation}
\renewcommand\thetable{Table \oldthetable}

\newlist{assumptions}{enumerate}{3}
\setlist[assumptions]{label*=\arabic*.,ref=Assumption \arabic*}

\begin{document}
\pagenumbering{arabic}
\title{Reduction of rain-induced errors for wind speed estimation on SAR observations using convolutional neural networks}

\author{Aurélien~Colin$^{1,2}$,
        Pierre~Tandeo$^{1,3}$,
        Charles~Peureux$^{2}$,
        Romain~Husson$^{2}$,
        Ronan~Fablet$^{1,3}$,
\thanks{$^1$ IMT Atlantique, Lab-STICC, UMR CNRS 6285, F-29238, France.}
\thanks{$^2$ Collecte Localisation Satellites, Brest, France.}
\thanks{$^3$ Odyssey, Inria/IMT, France}}

\markboth{Journal of \LaTeX\ Class Files,~Vol.~13, No.~9, September~2014}%
{Shell \MakeLowercase{\textit{et al.}}: Bare Demo of IEEEtran.cls for Journals}

\maketitle

\begin{abstract}
Synthetic Aperture Radar is known to be able to provide high-resolution estimates of surface wind speed. These estimates usually rely on a Geophysical Model Function (GMF) that has difficulties accounting for non-wind processes such as rain events. Convolutional neural network, on the other hand, have the capacity to use contextual information and have demonstrated their ability to delimit rainfall areas. By carefully building a large dataset of SAR observations from the Copernicus Sentinel-1 mission, collocated with both GMF and atmospheric model wind speeds as well as rainfall estimates, we were able to train a wind speed estimator with reduced errors under rain. Collocations with in-situ wind speed measurements from buoys show a root mean square error that is reduced by 27\% (resp. 45\%) under rainfall estimated at more than 1 mm/h (resp. 3 mm/h). These results demonstrate the capacity of deep learning models to correct rain-related errors in SAR products.
\end{abstract}


\IEEEpeerreviewmaketitle

\section{Introduction}

\IEEEPARstart{S}{ynthetic} Aperture Radar (SAR) is a powerful tool for studying the ocean surface. C-Band SAR are sensitive to variations in sea surface roughness, and have been used to detect various meteorological and ocean processes, referred to as metocean, such as atmospheric or ocean fronts \cite{10.1016/j.rse.2019.111457}, icebergs \cite{10.1016/j.isprsjprs.2019.08.015}, oil surfactants from pollution \cite{10.1109/igarss39084.2020.9323590} or generated by plankton \cite{10.1080/014311601450040}, and some species of seaweed \cite{10.1016/j.marpolbul.2013.10.044}. They are particularly useful for studying waves \cite{10.1029/2008gl037030} and extreme events like cyclones \cite{10.1175/bams-d-11-00001.1, 10.1109/tgrs.2017.2732508}. There has been particular attention given to estimating wind speed using these sensors.

As the number of satellite missions with C-SAR sensors increases and archives of these data accumulate, it is becoming easier to build large SAR datasets. This paper focuses on the Sentinel-1 mission from the Copernicus program, which consists of two satellites, Sentinel-1A (launched in 2014) and Sentinel-1B (launched in 2016, which has been out of operation since December 2021). Sentinel-1C is planned to be launched in 2023. Ground Range Detected Higher Resolution Interferometric Wide-swath (GRDH IW) observations have a range of 250 km, an azimuth of about 200 km, and a resolution of 10 m/px. These observations are mainly routinely acquired over coastal areas.
Systematic processes are used to produce geophysical products from these observations, including wind speed estimates. Several Geophysical Model Functions (GMFs) have been developed for this purpose, including CMOD3 \cite{10.1163/156939395x00532}, CMOD4 \cite{10.1029/96jc02860}, CMOD5 \cite{10.21957/6GHPJQGW3}, CMOD5.N \cite{10.21957/mzcfm6jfl}, CMOD6 \cite{ 10.1117/12.2195727}, CMOD7 \cite{10.1109/jstars.2017.2681806} and C\_SARMOD2 \cite{10.23919/piers.2018.8598163}. 
These GMFs use the vertical-vertical polarization, which is Sentinel-1's default co-polarization in coastal areas. However, the co-polarization channel saturates at high wind speed \cite{10.1080/01431160802555879}. Therefore, the cross-polarization has also been used to estimate the wind speed \cite{10.1175/bams-d-11-00001.1}. \textcolor{black}{In particular, H14E \cite{10.1002/2014JC010439} was found to provide accurate wind speed measurements even in extreme events such as category 5 hurricanes \cite{10.1029/2019jc015056}.} The cross-polarization has the advantage of lower incidence and wind direction dependency \cite{10.1175/jtech-d-13-00006.1}. GMFs have also been developed for horizontal-horizontal polarization \cite{10.1109/lgrs.2020.2967811}. \textcolor{black}{Bayesian nonparametric wind speed estimation has also been proposed, offering the advantage of not relying on wind direction priors that are sensitive to meteorological processes with rapid temporal and spatial evolution \cite{10.1109/tgrs.2022.3188328}.}

However, these GMFs are sensitive to contamination from non-wind processes. In particular, rainfall can either increase or decrease sea surface roughness \cite{10.1016/j.rse.2016.10.015}, making it difficult to correct for its effects.

Deep learning models, particularly Convolutional Neural Networks (CNNs), have demonstrated their ability to detect rain signatures in SAR observations \cite{10.48550/arxiv.2207.07333}. These models are known to be able to tackle denoising \cite{10.1109/mgrs.2021.3070956} and inpainting \cite{10.1016/j.jag.2022.103019} tasks because they use contextual information to estimate the original signal. This paper is dedicated to estimating wind speed in rainy areas using a model that does not require an explicit rainfall prior and only uses the parameters available to GMFs.

In the first section, we present the SAR data used to train the model and the ancillary information available. The second section describes the methodology used to build the dataset, with special attention given to ensuring a balanced representation of rainfall observations. The final section presents the results on the training set and confirms them with in-situ measurements from buoys, demonstrating the model's ability to correct for rain overestimates.

\section{Dataset}

The SAR measurements used in this chapter come from 19,978 IW observations acquired globally between March  03\textsuperscript{rd}, 2018 and the February 23\textsuperscript{rd}, 2022, inclusive. Each of these observations covers approximately 44 000 km² and has a resolution of 100 m/px, downscaled from the GRDH products available at 10 m/px. \textcolor{black}{The radiometric information used as input is the Sea Surface Roughness (SSR) defined in \cite{10.1002/gdj3.73} as the normalized radar cross section $\sigma_0$ divided by the $\sigma_0$ of a wind of 10 m/s and a direction of 45° relative to the antenna look angle. The $\sigma_0$ of this neutral wind is given by the GMF CMOD5.N.}

Obtaining global information on rain that can be used in conjunction with SAR observations can be difficult. A previous study conducted using a global Sentinel-1 dataset found only 2,304 partial collocations with the satellite-based radar GPM/DPR \cite{10.5067/GPM/DPR/GPM/2A/07} out of 182\textcolor{black}{,}153 IW. "Partial collocations" refers to instances where at least 20x20 km of a swath is observed by the spaceborne weather radar 20 minutes before or after the SAR observation. Coastal ground-based radars like NEXRAD \cite{10.7289/V5W9574V} could provide rainfall estimates, but they are affected by topography and may not capture all wind regimes. Therefore, SAR-based rain estimation is preferred to maximize the number of available observations and simplify the collocation process. We used a recent SAR rainfall estimator \cite{10.48550/arxiv.2207.07333} that emulates NEXRAD's reflectivity and proposes three rainfall thresholds that roughly correspond to 1 mm/h, 3 mm/h, and 10 mm/h.

Ancillary information, such as incidence angle and satellite heading, is retrieved from Sentinel-1 Level-2 products. It also includes collocations with atmospheric models from the European Centre for Medium-Range Weather Forecasts, which provide modelled wind speed and direction, as well as the surface wind speed computed by the GMF. The atmospheric models have a spatial resolution of 0.25x0.25 degrees and a temporal resolution of 3 hours \textcolor{black}{\cite{esaSentinel1Auxiliary}}, while the GMF \textcolor{black}{is computed at} a spatial resolution of 1 km/px and corresponds to the observation itself. \textcolor{black}{The GMF used in Level-2 product is IFR2 \cite{10.1029/97jc01911} until July 2019 and CMOD5.N \cite{10.21957/mzcfm6jfl} afterwards.} \textcolor{black}{The ancillary information contained in the Level-2 products is available at a spatial resolution of 1 km/px. However, it is interpolated to 100 m/px to match the grid of the radiometric channels.}

\section{Methodology}

This section presents the methodology for building the rain-invariant wind speed estimator. We first describe the deep learning architecture of the model, then we discuss the creation of the dataset, which is biased to have a large number of rain examples. The final section describes the evaluation procedure.

\subsection{Deep Learning Model}

\begin{figure}
    \centering
    \includegraphics[width=0.95\linewidth]{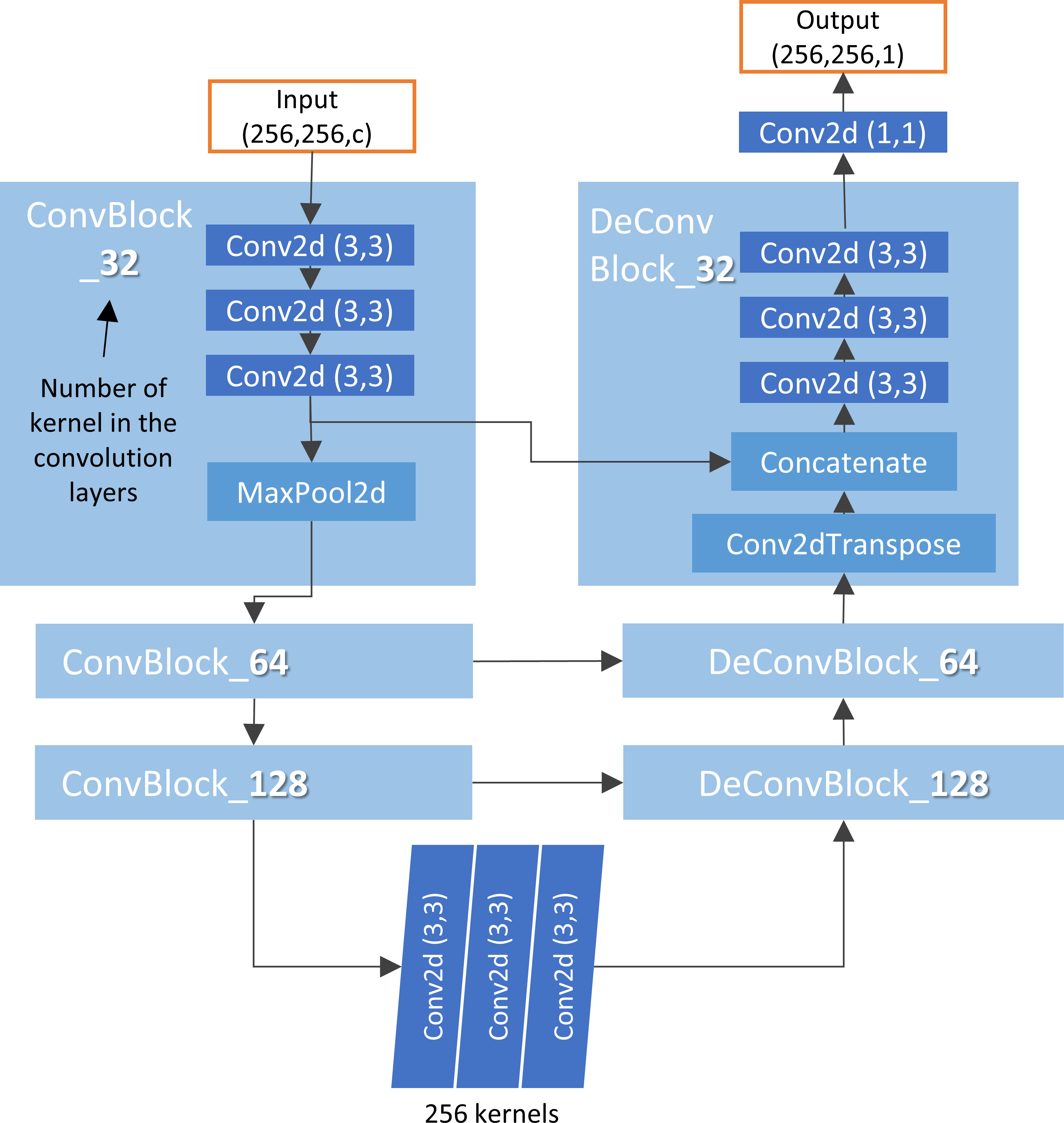}
    \caption{Architecture of the UNet model used for estimating the wind speed.}
    \label{fig:unet}
\end{figure}

The architecture used in this chapter is the UNet architecture \cite{DBLP:journals/corr/RonnebergerFB15} depicted in \ref{fig:unet}. UNet is an autoencoder architecture with the advantage of being fully convolutional, meaning it has translation equivariance properties (translations of the input result in translations of the output). In addition, skip connections between the encoder and the decoder facilitate training, especially by reducing the vanishing gradient issue \cite{10.48550/arxiv.1211.5063}). Introduced in 2015, UNet has been used in various domains and has demonstrated its importance for segmentation of SAR observations \cite{10.3390/rs14040851, 10.48550/arxiv.2207.07333, 10.1109/jstars.2021.3074068}.

The output of the model always contains a single convolution kernel, activated by the Rectified Linear Unit (ReLU) function to ensure that the prediction is in the interval [0, +$\infty$\textcolor{black}{]}. All convolution kernels in the hidden layers are also activated by ReLU functions. The model is set to take input of 256x256 pixels during training, but since the weights only describe convolution kernels, it is possible to use the model for inference on images of any shape as long as the input resolution remains at 100 m/px. Variants of the model are trained with different numbers of input channels. The architecture is modified by changing the size of the first convolution kernel, which is defined as a kernel of size (3, 3, c, 32), where c is the number of input channels.

\subsection{Dataset balancing procedure}

In this section, we describe the process for building a balanced dataset. Our goals are to (I) ensure that the wind distribution of each dataset is close to the real-world distribution, (II) prevent information leak between the training, validation, and test subsets, (III) ensure that the groundtruth wind speed, obtained from an atmospheric model, accurately represents the real-world wind speed, and (IV) include enough rain samples to allow the model to learn from them.

\paragraph{Rain and rainless patches selection}

As discussed earlier, rainfall estimation is provided by a deep learning model at a resolution of 100 m/px, on the same grid as the SAR observation. Therefore, it is possible to separate the observations into two areas, $\mathcal{A}^+$ and $\mathcal{A}^-$, based on the 3 mm/h threshold from the rainfall estimation.

\begin{align}
    \mathcal{A}^+ = \{x: Rainfall(x) >= 3 mm/h\} \\
    \mathcal{A}^- = \{x: Rainfall(x) < 3 mm/h\}
\end{align}

However, most SAR observations do not contain rain signatures. Collocations with GPM's dual polarization radar, a satellite-based weather radar, indicated that the probability of rain rates higher than 3 mm/h was 0.5\%. Thus, by dividing the SAR observations into tiles of 256 by 256 pixels, we call "rain patches" the tiles with more than 5\% of their surface predicted to have rain rates higher than 3 mm/h, and "rainless patches" those without rain signatures. We denote $n_+$ as the number of rain patches and $n_-$ as the number of rainless patches. To ensure that the model will learn regardless of the rain-situation, we set $n_+ = n_-$.

\begin{figure}[ht!]
    \centering
    
    (a)
    
    \includegraphics[width=0.95\linewidth]{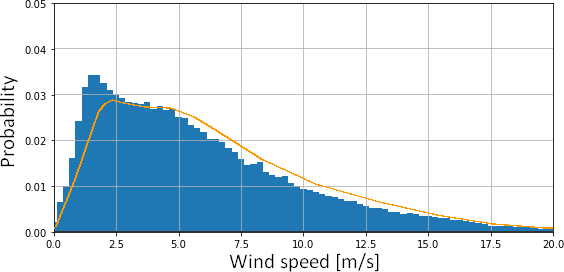} 
    
    (b)
    
    \includegraphics[width=0.95\linewidth]{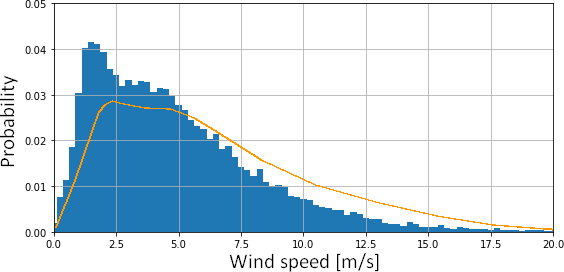}
    \caption{Distribution of ERA5 wind speed collocated with surface rain rate from GPM/DPR for rainfall higher than 3 mm/h (a) and 30 mm/h (b). \textcolor{black}{They amount for 0.55\% and 0.02\% of the collocations, respectively. The orange curve in the figure shows the wind distribution regardless of the rainfall.}}
    \label{fig:wind_speed_estimation/GPMvsECMWF}
\end{figure}

\ref{fig:wind_speed_estimation/GPMvsECMWF} indicates collocations between the reanalysis from ERA5 and the satellite-based weather radar GPM/DPR. It shows the wind speed distribution for rainfall higher than 3 mm/h (a), 30 mm/h (b), and the overall distribution (orange curve). Stronger rainfall is associated with a lower probability of strong wind speeds. However, the impact is mostly marginal at moderate rain rates.

\paragraph{Restriction to \textit{a priori} Accurate Model Wind Speeds}

Atmospheric models have been known to have \textcolor{black}{coarse} resolution (0.25x0.25 degrees spatially, 3 hours temporally) and to be unable to accurately depict fine-scale wind fields. However, they are computed globally and independently of the SAR observations. On the other hand, SAR-based wind fields from the GMFs are known to be accurate on rainless patches, \textcolor{black}{but are diversely affected by rainfall. At low and moderate wind speeds, impinging droplets lead to an overestimation of the wind speed. Under high wind speed conditions, dominant attenuation results in an underestimation of the wind speed. Because of the difficulty of collocating a sufficient quantity of high wind speed and rainfall events, this study mainly aims to correct the wind speed overestimation.} We calculate $\Delta_{\mathcal{A^-}}$ as the discrepancy between the GMF and the atmospheric model on rain-free pixels.

\begin{equation}
    \Delta_{\mathcal{A^-}} = MSE_{|\mathcal{A^-}}(Atm, GMF)
\end{equation}

In our experiments, the threshold was set at $\Delta_{\mathcal{A}^-} < 1$ m/s. All patches containing a higher discrepancy between the two wind speed sources were discarded. \textcolor{black}{By ensuring agreement between the atmospheric model and the SAR-based GMF on rainless patches, we assume that the atmospheric model is close to the real wind speed and use it as a target for optimizing the deep learning model.}

\paragraph{Balancing to the real-world wind distribution}

It should be noted that this condition ensures accurate modeled wind speeds and rain distribution, especially because the rainfall estimator is known to overestimate rainfall at high wind speeds.

We denote:
\begin{itemize}
    \item $P^+$ as the wind speed distribution on $n_+$.
    \item $P^-$ as the wind speed distribution on $n_-$.
    \item $P$ as the wind speed distribution on $n_- \cup n_+$.
\end{itemize}

Balancing the dataset to the real-world wind distribution translates to the following condition:

\begin{equation}
    \forall x, P(x) = \frac{n_+P^+(x)+n_-P^-(x)}{n_++n_-}
\label{eq:dataset_policy}
\end{equation}

As we choose to keep all rain patches and to set $n_+ = n_-$, \ref{eq:dataset_policy} leads to:

\begin{equation}
    \forall x, P^-(x) = \frac{1}{2}(P^+(x)-P(x))
\label{eq:dataset_policy2}
\end{equation}

\begin{figure}
    \centering
    \includegraphics[width=0.95\linewidth]{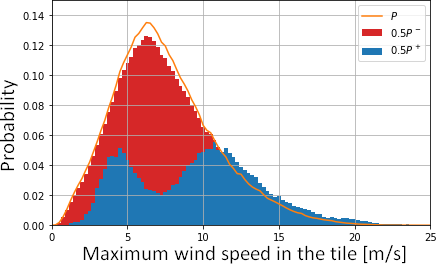}
    \caption{Wind speed distributions for rain (blue) and rainless (red) patches. Orange curve correspond to the real world wind distribution.}
    \label{fig:dataset_wind_distribution}
\end{figure}

For some wind speeds $x$, $P^+(x)$ is higher than twice $P(x)$. In these cases, we relax the condition from \ref{eq:dataset_policy2} in order to avoid removing rain patches. \ref{fig:dataset_wind_distribution} depicts the wind speed distribution for rain and rainless patches. The mean squared error between $P$ and $\frac{1}{2}(P^++P^-)$ reach 8.8\%.

The dataset can be further balanced to ensure that, for each wind speed, the number of rain and rainless patches is equal. However, this leads to remove of 84\% of the data. Appendix 1 compares the performance of this second dataset. As did not provide improvements, this dataset is left out of the main document.

\subsubsection{Training, Validation and Test Set Division}
\label{section:Training}

After extracting the patches following the distributions $P^+$ and $P^-$, they are split into training, validation, and test sets. Each subset preserves the same distributions. Furthermore, to avoid information leakage, if a patch from one IW is in a subset, every patch from the same IW belongs to the same subset. The stochastic brute forcing method described in Algorithm
\ref{alg:stochastic_brute_forcing} draws random IWs and computes the distribution of the validation and test subsets, compares them to the overall distributions, and returns the solution that minimizes the difference. In this algorithm, $\bar{P}_e$ indicates the wind speed distribution multiplied by the number of patches in $e$ and divided by the total number of patches. It ensures that the validation and test subsets each contain approximately 10\% of all the patches.

\begin{algorithm}[ht!]
\caption{Stochastic brute forcing}
\small
\label{alg:stochastic_brute_forcing}
    \SetKwInOut{Input}{input}
    \SetKwInOut{Output}{output}
    \Input{The list $L$ of all swath $i$ and their patch distribution $P_i$.}
    \Output{$L_{val}$ and $L_{test}$, the list of the swaths contained in the validation and test sets.}
    \bigskip
    $\mathcal{L}$ = MAE\;
    $e_{min} = +\infty$\;
    $n_1$ = len(L)/10\;
    $n_2$ = len(L)/3\;
    \For{$i\gets0$ \KwTo $1000000$ by $1$}{
        $n$ = random.integer(min=$n_1$, max=$n_2$)\;
        $candidate$ = random.choice($L$, size=$n$, replace=False)\;
        $c_{val}$ = $candidate[:n/2]$\;
        $c_{test}$ = $candidate[n/2:]$\;
        $e_{val} = \mathcal{L}(0.1P^+, \bar{P}^-_{c_{val}})+\mathcal{L}(0.1P^-, \bar{P}^-_{c_{val}})$\;
        $e_{test} = \mathcal{L}(0.1P^+, \bar{P}^+_{c_{test}})+\mathcal{L}(P^-, 0.\bar{P}P^-_{c_{test}})$\;
        
        $e = \frac{2\cdot e_{val} \cdot\; e_{e_test}}{e_{val} + e_{test}}$\;
        \If{$e < e_{min}$}{
            $L_{val} = c_{val}$\;
            $L_{test} = c_{test}$\;
            $e_{min} = e$\;
            }
        }
\end{algorithm}

\textcolor{black}{The initial 19,978 IW observations account for 105,164 rain patches and 2,094,370 rainless patches. At the end of the process, the distribution in the subsets is as follow:}

\begin{itemize}
    \item \textcolor{black}{168,349 patches from 14,169 IWs in the training set;}
    \item \textcolor{black}{20,944 patches from 1,763 IWs in the test set;}
    \item \textcolor{black}{21,010 patches from 1,763 IWs in the validation set.}
\end{itemize}

Before training the model, we compute the mean and standard deviation of each channel on the training set and use them to normalize the inputs during training, validation, and inference. The output, however, is not normalized. We train the model for 100,000 weight updates \textcolor{black}{(i.e. steps of the stochastic gradient descent)} to minimize the Mean Square Error (MSE) \textcolor{black}{between the prediction and the wind speed from the atmospheric model}, with a batch size of 16 and a learning rate of $10^{-5}$ using the Adam optimizer.

\subsection{Evaluation procedure}

To evaluate the impact of each input channel, we train various variants of the model:

\begin{itemize}
        \item \Romannum{1} uses only the VV channel.
        \item \textcolor{black}{\Romannum{2} uses the VV channel, the incidence angle and the \textit{a priori} wind direction.}
        \item \textcolor{black}{\Romannum{3} uses both the VV and the VH channel, the incidence angle and the \textit{a priori} wind direction.}
        \item \textcolor{black}{\Romannum{4} uses both the VV and the VH channel, the incidence angle, the \textit{a priori} wind direction and the wind speed prior.}
        \item \textcolor{black}{\Romannum{5} uses only the wind speed prior.}
\end{itemize}
\textcolor{black}{All channels are interpolated to 100 m/px and concatenated on the same grid.} 

Incidence angles and wind directions from European Centre for Medium-Range Weather Forecasts (ECMWF) atmospheric model are obtained from the Level-2 products and notably used for the computation of the GMF. Therefore, architecture \Romannum{2} contains the same inputs as the GMF \textcolor{black}{though interpolated at a finer resolution.}

Each architecture is trained five times to reduce the impact of random initialization on the evaluation results. The results is presented as the mean and standard deviation over these five independent trainings.

We compare the results using the Root Mean Square Error (RMSE) and the Pearson correlation coefficient (PCC). The PCC is formulated in \ref{eq:PCC}.

\begin{equation}
    PCC_{Y, \hat{Y}} = \frac{\mathds{E}[(Y - \mu_Y)(\hat{Y} - \mu_{\hat{Y}})]}{\sigma_Y \sigma_{\hat{Y}}} \label{eq:PCC}
\end{equation}

The results are computed against both the groundtruths from the atmospheric model, which provides a large test set, and against collocations with buoys, which have good temporal resolution and are in-situ measurements.

\section{Results}

\subsection{Benchmarking experiments}

The performance of the models compared to ECMWF are calculated on the test subset for each input variant and the baseline GMF. The results of this analysis can be found in \ref{table:wind_speed_results}. It appears that the most important input is the GMF itself, as both \Romannum{4} and \Romannum{5} have better results than the other variants. \Romannum{1}, \Romannum{2} and \Romannum{3} are unable to achieve better results than the GMF, except under strong rainfall, even though \Romannum{2} and \Romannum{3} have access to all the channels used by the GMF.

\begin{table*}[!ht]
    \scriptsize
    \centering
    \begin{tabularx}{1.\textwidth}{|YY|Y;{1pt/2pt}Y|Y;{1pt/2pt}Y|}\hline
    & & \multicolumn{2}{c|}{ RMSE} & \multicolumn{2}{c|}{ PCC}\\
    & & & & & \\
    \multirowcell{-3}{MODEL AND\\CHANNELS} & \multirowcell{-3}{RAIN RATE} & \multirowcell{-2}{Balanced\\Dataset} & \multirowcell{-2}{Neutral\\Dataset} & \multirowcell{-2}{Balanced\\Dataset} & \multirowcell{-2}{Neutral\\Dataset} \\\hline
    
    & $[0,1[ mm/h$ & 
    \textcolor{white}{$\Diamond$} 1.38 [0.016] & 
    \textcolor{white}{$\Diamond$}\bf 2.40 [0.019]  & 
    \textcolor{white}{$\Diamond$} 89.9\% [0.19\%] & 
    \textcolor{white}{$\Diamond$} 74.2\% [0.42\%]  \\
    & $[1,3[ mm/h $ & 
    \textcolor{white}{$\Diamond$} 1.64 [0.046] & 
    \textcolor{white}{$\Diamond$}\bf 2.78 [0.038]  & 
    \textcolor{white}{$\Diamond$} 92.7\% [0.39\%] & 
    \textcolor{white}{$\Diamond$} 79.5\% [0.58\%]  \\
    & $[3,10[ mm/h$ & 
    \textcolor{white}{$\Diamond$} 1.59 [0.038] & 
    \textcolor{white}{$\Diamond$}\bf 3.18 [0.070]  & 
    \textcolor{white}{$\Diamond$}\bf 92.7\% [0.38\%] & 
    \textcolor{white}{$\Diamond$} 78.9\% [0.62\%]  \\
    \multirowcell{-4}{\Romannum{1}\\ $[$VV$]$}
    & $\geq 10 mm/h$ & 
    \textcolor{white}{$\Diamond$}\bf 2.12 [0.052] & 
    \textcolor{black}{$\bigstar$}\bf 3.27 [0.135]  & 
    \textcolor{white}{$\Diamond$}\bf 81.7\% [1.03\%] & 
    \textcolor{white}{$\Diamond$} 74.3\% [0.90\%] \\ \hline
    
    & $[0,1[ mm/h$ & 
    \textcolor{white}{$\Diamond$} 0.87 [0.009] & 
    \textcolor{white}{$\Diamond$}\bf 2.26 [0.019]  & 
    \textcolor{white}{$\Diamond$} 96.2\% [0.06\%] & 
    \textcolor{white}{$\Diamond$} 77.7\% [0.24\%]  \\
    & $[1,3[ mm/h$ & 
    \textcolor{white}{$\Diamond$} 1.04 [0.086] & 
    \textcolor{white}{$\Diamond$}\bf 2.87 [0.061]  & 
    \textcolor{white}{$\Diamond$} 97.2\% [0.47\%] & 
    \textcolor{white}{$\Diamond$} 77.6\% [0.24\%]  \\
    & $[3,10[ mm/h$ & 
    \textcolor{white}{$\Diamond$}\bf 1.19 [0.053] & 
    \textcolor{white}{$\Diamond$}\bf 3.39 [0.193]  & 
    \textcolor{white}{$\Diamond$} 96.2\% [0.37\%] & 
    \textcolor{white}{$\Diamond$} 75.0\% [3.04\%]  \\
    \multirowcell{-4}{\Romannum{2}\\ $[$VV, INC, WDIR$]$}
    & $\geq 10 mm/h$ & 
    \textcolor{white}{$\Diamond$}\bf 2.27 [0.151] & 
    \textcolor{white}{$\Diamond$} 3.86 [0.700]  & 
    \textcolor{white}{$\Diamond$}\bf 81.7\% [1.89\%] & 
    \textcolor{white}{$\Diamond$} 63.2\% [9.19\%]  \\ \hline
    
    & $[0,1[ mm/h$ & 
    \textcolor{white}{$\Diamond$} 0.83 [0.002] & 
    \textcolor{white}{$\Diamond$} 2.18 [0.006] & 
    \textcolor{white}{$\Diamond$} 96.5\% [0.02\%] & 
    \textcolor{white}{$\Diamond$} 79.3\% [0.07\%]  \\
    & $[1,3[ mm/h$ & 
    \textcolor{white}{$\Diamond$} 0.93 [0.020] & 
    \textcolor{white}{$\Diamond$}\bf 2.73 [0.022]  & 
    \textcolor{white}{$\Diamond$} 97.7\% [0.07\%] & 
    \textcolor{white}{$\Diamond$}\bf 80.1\% [0.29\%]  \\
    & $[3,10[ mm/h$ & 
    \textcolor{white}{$\Diamond$}\bf 1.09 [0.022] & 
    \textcolor{white}{$\Diamond$}\bf 3.25 [0.052]  & 
    \textcolor{white}{$\Diamond$}\bf 96.7\% [0.07\%] & 
    \textcolor{white}{$\Diamond$} 78.0\% [0.44\%]  \\
    
    \multirowcell{-4}{\Romannum{3}\\ $[$VV, VH, \\INC, WDIR$]$}
    &  $ \geq 10 mm/h$ & 
    \textcolor{white}{$\Diamond$}\bf 2.13 [0.050] & 
    \textcolor{white}{$\Diamond$}\bf 3.68 [0.310]  & 
    \textcolor{white}{$\Diamond$}\bf 83.9\% [0.66\%] & 
    \textcolor{white}{$\Diamond$} 70.8\% [2.95\%]  \\ \hline
    
    & $[0,1[ mm/h$ & 
    \textcolor{black}{$\bigstar$}\bf 0.64 [0.007] & 
    \textcolor{black}{$\bigstar$}\bf 1.90 [0.028] & 
    \textcolor{black}{$\bigstar$}\bf 97.9\% [0.04\%] & 
    \textcolor{black}{$\bigstar$}\bf 85.0\% [0.33\%] \\
    & $[1,3[ mm/h$ & 
    \textcolor{black}{$\bigstar$}\bf 0.63 [0.015] & 
    \textcolor{black}{$\bigstar$}\bf 2.29 [0.075]  & 
    \textcolor{black}{$\bigstar$}\bf 98.9\% [0.03\%] & 
    \textcolor{black}{$\bigstar$}\bf 87.1\% [0.56\%]  \\
    & $[3,10[ mm/h$ & 
    \textcolor{black}{$\bigstar$}\bf 0.78 [0.040] & 
    \textcolor{black}{$\bigstar$}\bf 2.55 [0.132] & 
    \textcolor{black}{$\bigstar$}\bf 98.4\% [0.08\%] & 
    \textcolor{black}{$\bigstar$}\bf 87.1\% [1.12\%] \\
    \multirowcell{-4}{\Romannum{4}\\ $[$VV, VH,\\INC, WDIR, GMF$]$}
    & $\geq 10 mm/h$ & 
    \textcolor{black}{$\bigstar$}\bf 1.63 [0.162] & 
    \textcolor{white}{$\Diamond$}\bf 3.37 [0.113] & 
    \textcolor{black}{$\bigstar$}\bf 90.9\% [1.17\%] & 
    \textcolor{white}{$\Diamond$} 73.4\% [2.31\%] \\ \hline
    
    & $[0,1[ mm/h$ & 
    \textcolor{white}{$\Diamond$}\bf 0.67 [0.003] & 
    \textcolor{white}{$\Diamond$} 3.16 [2.650] & 
    \textcolor{white}{$\Diamond$}\bf 97.7\% [0.02\%] & 
    \textcolor{white}{$\Diamond$} 80.7\% [17.55\%] \\
    & $[1,3[ mm/h$ & 
    \textcolor{white}{$\Diamond$}\bf 0.68 [0.005] & 
    \textcolor{white}{$\Diamond$} 4.17 [3.582] & 
    \textcolor{white}{$\Diamond$}\bf 98.8\% [0.01\%] & 
    \textcolor{white}{$\Diamond$}\bf 82.5\% [12.67\%] \\
    & $[3,10[ mm/h$ & 
    \textcolor{white}{$\Diamond$}\bf 0.88 [0.021] & 
    \textcolor{white}{$\Diamond$} 4.53 [3.739]  & 
    \textcolor{white}{$\Diamond$}\bf 98.0\% [0.03\%] & 
    \textcolor{white}{$\Diamond$}\bf 83.4\% [8.19\%]  \\
    \multirowcell{-4}{\Romannum{5}\\$[$GMF$]$}
    &  $\geq 10 mm/h$ & 
    \textcolor{white}{$\Diamond$}\bf 1.94 [0.087] & 
    \textcolor{white}{$\Diamond$} 4.79 [2.917] & 
    \textcolor{white}{$\Diamond$}\bf 87.9\% [0.66\%] & 
    \textcolor{white}{$\Diamond$} 74.0\% [8.19\%]  \\ \hline
    
    & $[0,1[ mm/h$ & 
    \textcolor{white}{$\Diamond$} 0.77 & 
    \textcolor{white}{$\Diamond$} 2.41 & 
    \textcolor{white}{$\Diamond$} 97.0\% & 
    \textcolor{white}{$\Diamond$} 81.2\% \\
    
    & $[1,3[ mm/h$ & 
    \textcolor{white}{$\Diamond$} 0.84 & 
    \textcolor{white}{$\Diamond$} 3.16 & 
    \textcolor{white}{$\Diamond$} 98.1\% & 
    \textcolor{white}{$\Diamond$} 80.0\%  \\
    
    & $[3,10[ mm/h$ & 
    \textcolor{white}{$\Diamond$} 1.25 & 
    \textcolor{white}{$\Diamond$} 3.42 & 
    \textcolor{white}{$\Diamond$} 96.5\% & 
    \textcolor{white}{$\Diamond$} 81.9\% \\
    
    \multirowcell{-4}{ GMF} 
    &  $\geq 10 mm/h$ & 
    \textcolor{white}{$\Diamond$} 4.65 & 
    \textcolor{white}{$\Diamond$} 3.70 & 
    \textcolor{white}{$\Diamond$} 52.5\% & 
    \textcolor{black}{$\bigstar$} 75.4\% \\ \hline
    
    \end{tabularx}
    \caption{Comparison of the five variants of the model and the two datasets. RMSE and PCC are computed on the respective test set and for five training with random initialization. Results are given as mean and standard deviation in brackets. The best result for each metric is indicated by $\bigstar$. Results better than the GMF are in bold.}
    \label{table:wind_speed_results}
\end{table*}

\begin{figure}[ht!]
    \centering
    \begin{tabular}{cc}
        (a) & (b) \\
        \includegraphics[width=0.45\linewidth]{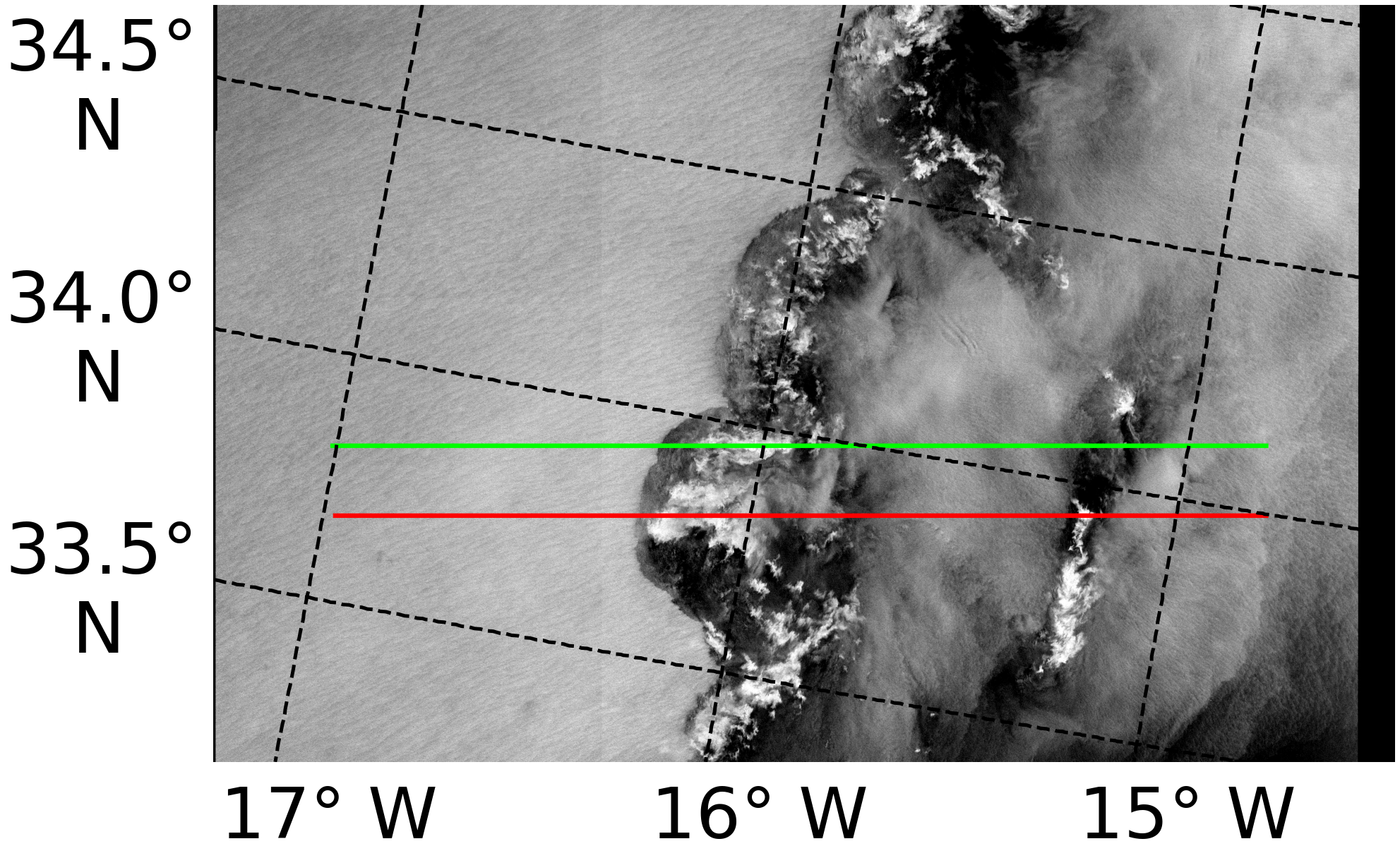} 
        & \includegraphics[width=0.45\linewidth]{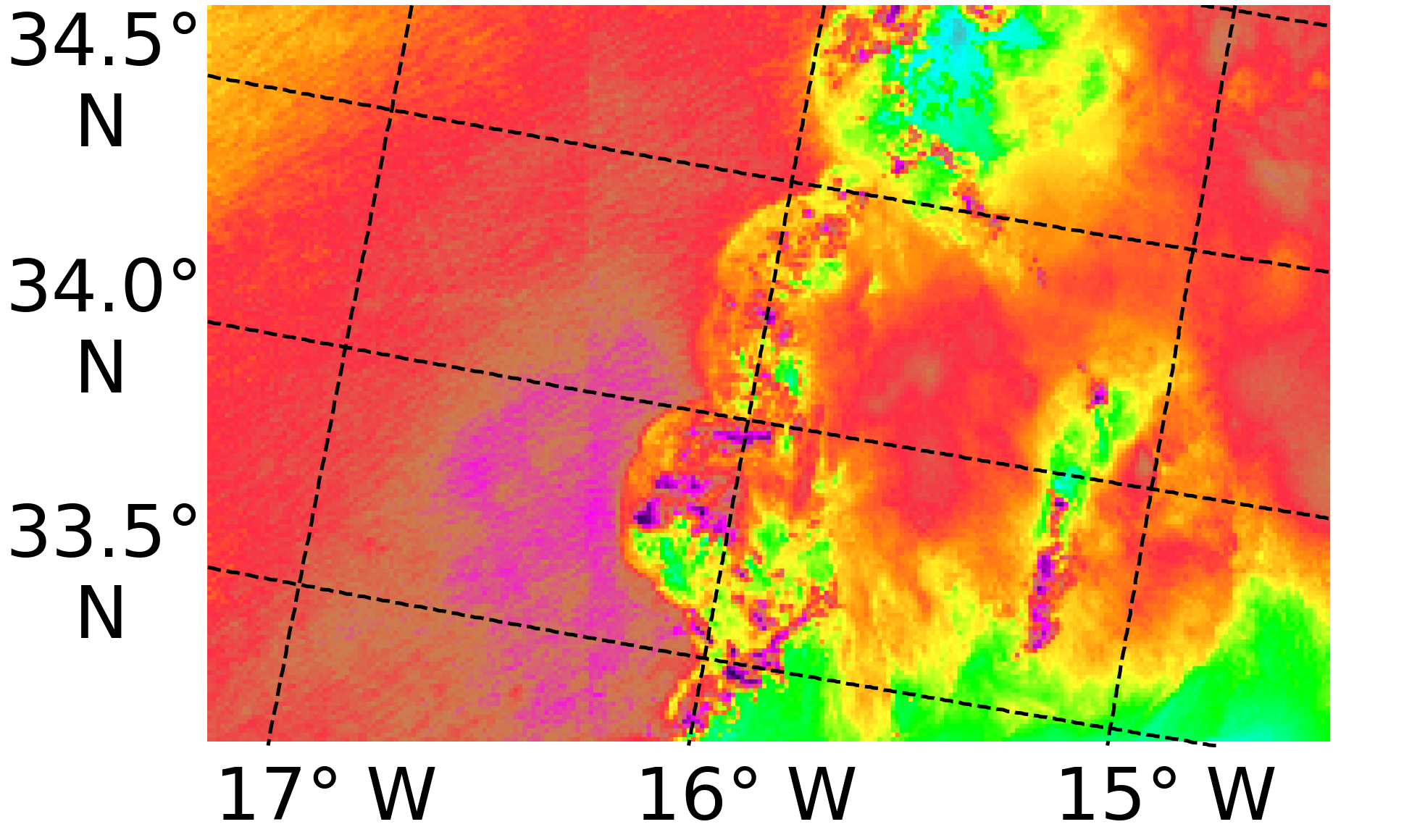} \\
        \includegraphics[width=0.45\linewidth]{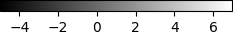}
        & \includegraphics[width=0.45\linewidth]{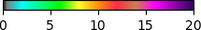}  \\
        (c) & (d) \\
        \includegraphics[width=0.45\linewidth]{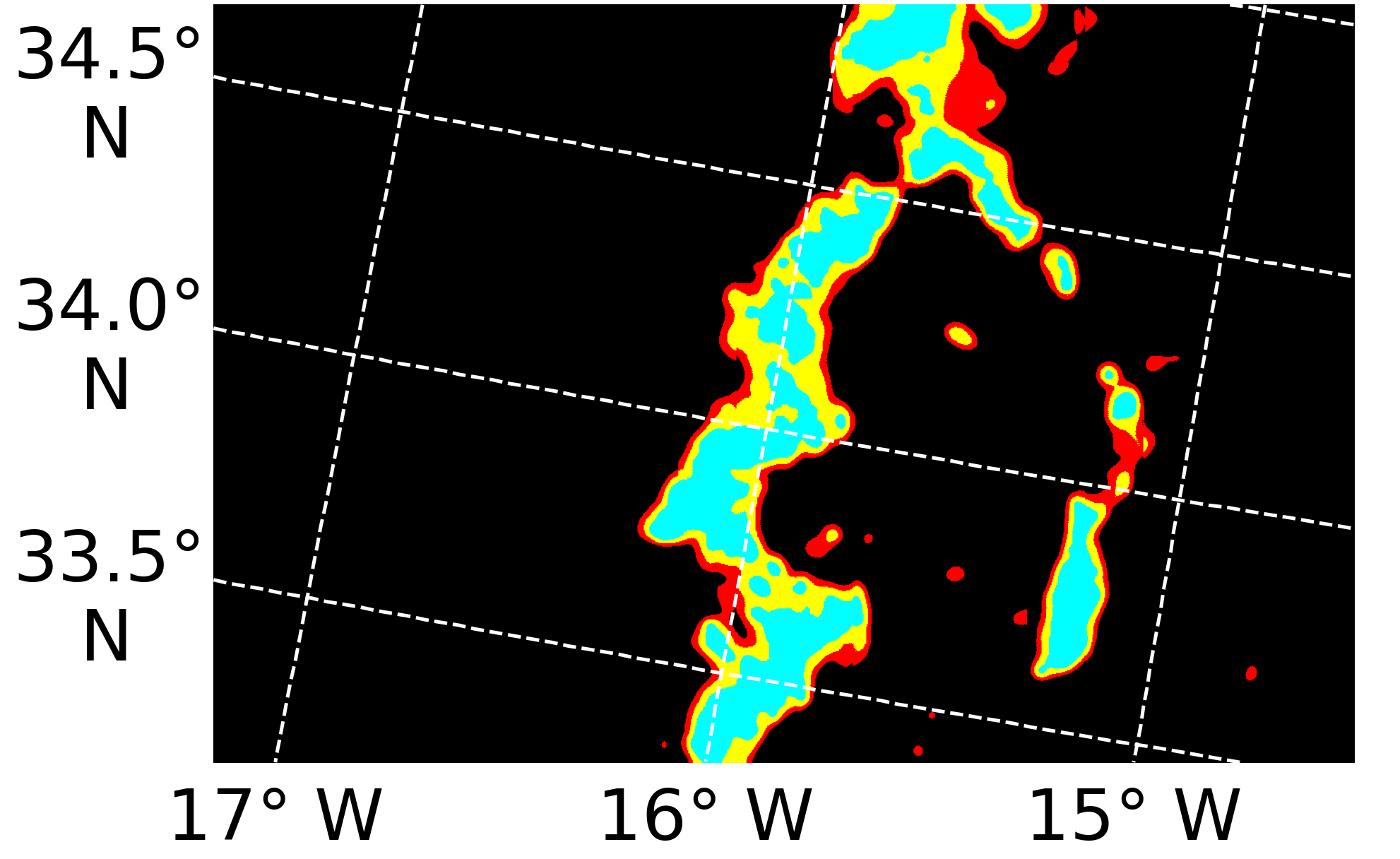} 
        & \includegraphics[width=0.45\linewidth]{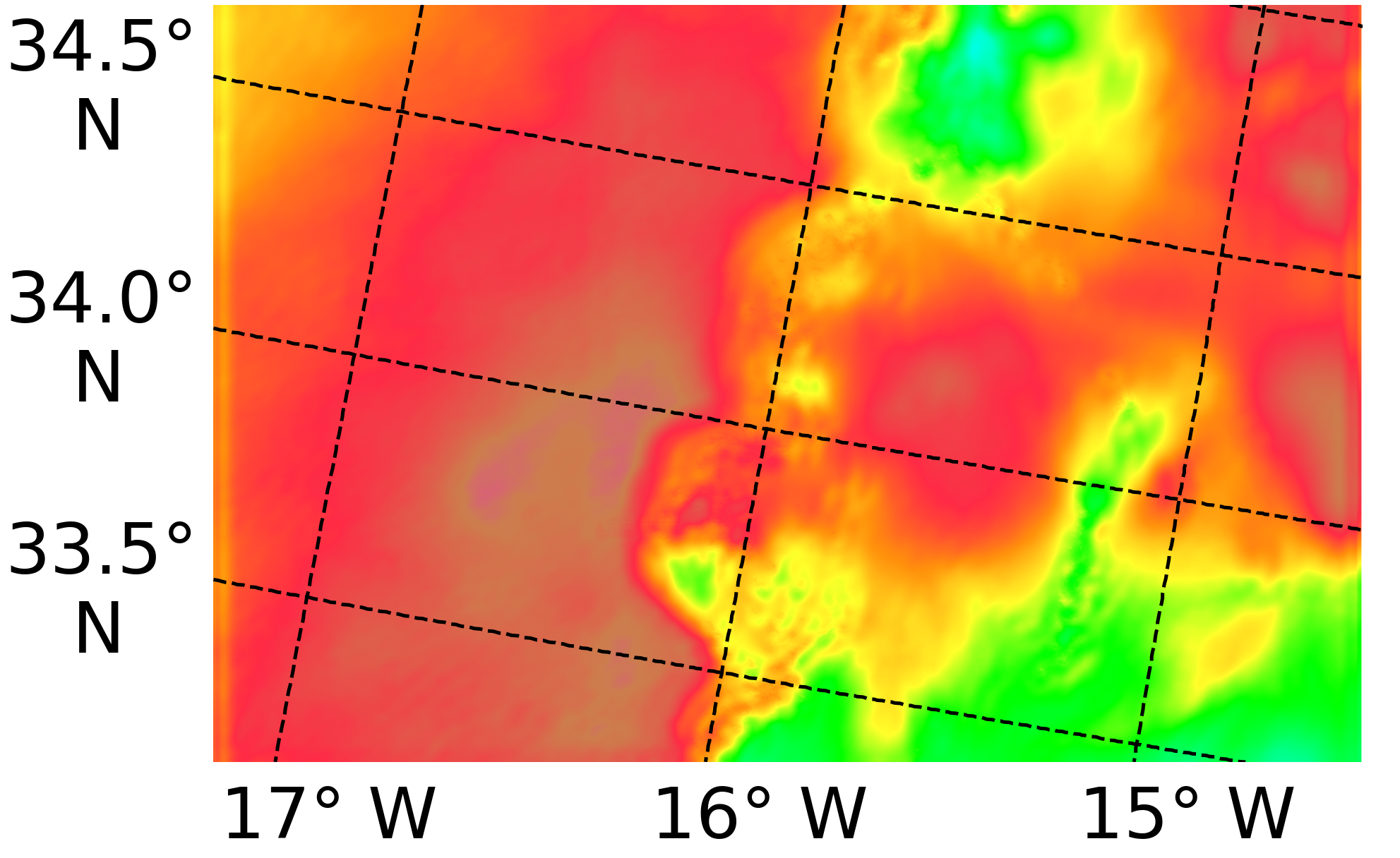} \\
        \includegraphics[width=0.45\linewidth]{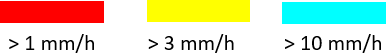}
        & \includegraphics[width=0.45\linewidth]{src/cmap/cmap_low_wind.png}  \\
        
        (e) & (f) \\
        \includegraphics[width=0.45\linewidth]{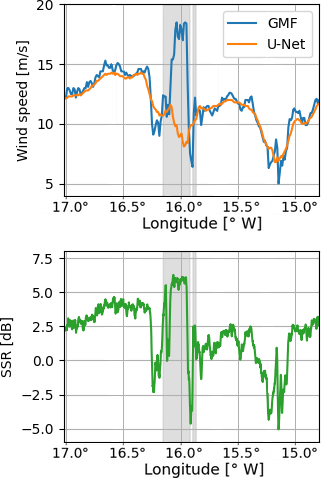} & 
        \includegraphics[width=0.45\linewidth]{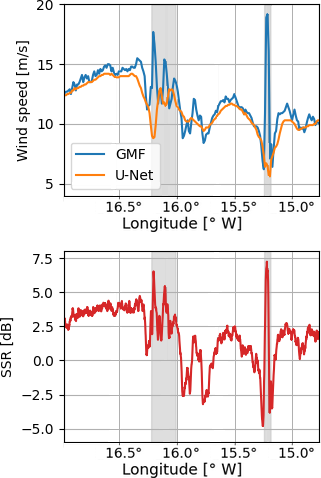} 
    \end{tabular}
    
    \caption{Observation from the November 11\textsuperscript{th} 2018 at 04:56:47. Sea Surface Roughness (SSR) of the VV channel in dB (a), geophysical model function estimated wind speed in m/s (b), deep learning estimated wind speed in m/s (d), \textcolor{black}{prediction of the rainfall estimator (c) and transects of the green (e) and red (f) lines, with rainfall predicted higher than 10 mm/h indicated as grey areas.}}
    \label{fig:deeplearning_wind_examples}
\end{figure}

Examples of rain-induced wind speed overestimation are presented in \ref{fig:deeplearning_wind_examples}. \ref{fig:deeplearning_wind_examples}.a shows the SSR on a logarithmic scale, with bright patterns appearing in the observation and segmented as rainfall in \ref{fig:deeplearning_wind_examples}.c. These patterns correlate with high wind speeds estimated by the GMF in \ref{fig:deeplearning_wind_examples}.b. The deep learning method in \ref{fig:deeplearning_wind_examples}.d ignores this spike in the SSR, but it also seems to underestimate the wind speed in the bottom-left quarter of the observation. Both of these behaviours are especially visible in the transect shown in \ref{fig:deeplearning_wind_examples}.f.

\subsection{Application to SAR observation with groundtruthed \textit{in-situ} data}

However, ECMWF wind speeds are \textcolor{black}{model} data and not in-situ, which can be obtained using anemometers on buoys. Using the dataset created in \cite{10.5194/wes-7-1441-2022}, 4732 collocation points between Sentinel-1 and National Data Buoy Center (NDBC) buoys are identified. The rain prediction model estimates that 4643 of these points are rainless, 75 record rainfall of more than 1 mm/h, and 14 record rainfall of more than 3 mm/h.
On a side note, the height at which in-situ measurements were taken varies, with most being between 3.8 m and 4.1 m above sea level. As mentioned in \cite{10.5194/wes-7-1441-2022}, the SAR inversion and deep learning prediction are both normalized to the altitude of the corresponding in-situ measurement. \textcolor{black}{Denoting $w$ the wind speed at 10 m estimated from the SAR, the wind speed at elevation $h$ is given by} the exponential law \cite{10.1175/1520-0450(1978)017<0390:otuopl>2.0.co;2}:

\begin{equation}
    \color{black} w_h = w \cdot \left(\frac{h}{10}\right)^{0.11}
\end{equation}

\ref{table:vs_buoy} indicates that the performances of the deep learning are higher than the GMF for both the Root Mean Square Error and the Pearson Correlation Coefficient for all rain ranges. The RMSE decreases by 0.04, 0.3\textcolor{black}{7} and 1.33 m/s for rainless, light rain, and moderate rain-situations, respectively. \textcolor{black}{The bias decreases by  0.26 and 1.01 m/s under light and moderate rain, but increases by 0.02 m/s for rainless situations.} \ref{table:vs_buoy} also demonstrates the importance of the dataset building scheme as a dataset composed of collocations without the aformentionned sample selection, referred to as the "neutral dataset," consistently has lower performances than the balanced dataset.

\begin{table}[!ht]
    \scriptsize
    \centering
    \resizebox{\linewidth}{!}{%
    \begin{tabular}{|cc||c|c|c|}\hline
        && \multirowcell{2}{Balanced\\Dataset} & \multirowcell{2}{Neutral\\Dataset} &GMF \\&&&&\\\hline
        
        \multirowcell{3}{\rotatebox[origin=c]{90}{Bias}}& $< 1 mm/h$ & 
        \textcolor{white}{$\Diamond$} 0.73 [0.04] &  
        \textcolor{white}{$\Diamond$} 1.32 [0.04]  & 
        \textcolor{black}{$\bigstar$} 0.71
        \\
         & $[1, 3] mm/h$ & 
        \textcolor{black}{$\bigstar$}\bf 1.38 [0.04] &  
        \textcolor{white}{$\Diamond$}\bf 1.47 [0.04]  & 
        \textcolor{white}{$\Diamond$} 1.64  
        \\
        & $> 3 mm/h$ & 
        \textcolor{black}{$\bigstar$}\bf 0.92 [0.07] &
        \textcolor{white}{$\Diamond$} 1.96 [0.29] &  
        \textcolor{white}{$\Diamond$} 2.\textcolor{black}{38}  
        \\\hline
        
        \multirowcell{3}{\rotatebox[origin=c]{90}{RMSE}} & $< 1 mm/h$ & 
        \textcolor{black}{$\bigstar$}\bf 1.44 [0.03] &  
        \textcolor{white}{$\Diamond$} 1.76 [0.12] & 
        \textcolor{white}{$\Diamond$} 1.48 
        \\
         & $[1, 3] mm/h$ & 
        \textcolor{black}{$\bigstar$}\bf 1.81 [0.04] &  
        \textcolor{white}{$\Diamond$}\bf 1.95 [0.18] & 
        \textcolor{white}{$\Diamond$} 2.18
        \\
        & $> 3 mm/h$ & 
        \textcolor{black}{$\bigstar$}\bf 1.60 [0.10] &  
        \textcolor{white}{$\Diamond$}\bf 2.42 [0.21] & 
        \textcolor{white}{$\Diamond$} 2.93  
        \\\hline
        
        \multirowcell{3}{\rotatebox[origin=c]{90}{PCC}} & $< 1 mm/h$ & 
        \textcolor{black}{$\bigstar$}\bf 93.6\% [0.16\%] &  
        \textcolor{white}{$\Diamond$} 92.9\% [0.20\%]  & 
        \textcolor{white}{$\Diamond$} 93.4\%  
        \\
         & $[1, 3] mm/h$ & 
        \textcolor{black}{$\bigstar$}\bf 96.3\% [0.22\%] &  
        \textcolor{white}{$\Diamond$} 93.4\% [0.19\%]  & 
        \textcolor{white}{$\Diamond$} 95.3\%  
        \\
         & $> 3 mm/h$ & 
        \textcolor{black}{$\bigstar$}\bf 95.9\% [0.35\%] &  
        \textcolor{white}{$\Diamond$}\bf 93.4\% [2.04\%] & 
        \textcolor{white}{$\Diamond$} 91.3\%  
        \\\hline

    \end{tabular}}
    \caption{Bias, Root Mean Square Error, and Pearson Correlation Coefficient of model \Romannum{4}, the GMF, for each rainfall level. The best result for each metric and rainfall level is indicated by $\bigstar$. Results better than the baseline are in bold. Results are given as mean and standard deviation in brackets.}
    \label{table:vs_buoy}
\end{table}

\begin{figure}
    \centering
    \begin{tabular}{cc}
        (a) & (b) \\
        \includegraphics[width=0.45\linewidth]{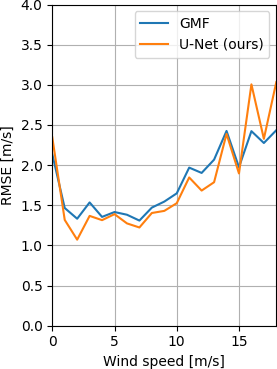} & \includegraphics[width=0.45\linewidth]{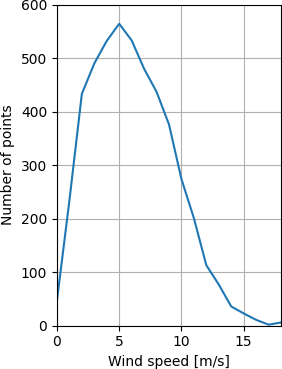}
    \end{tabular}
    \caption{\textcolor{black}{Evaluation of the Root Mean Square Error across the wind speed (a) and number of collocation with NDBC buoys in rainless areas (b).}}
    \label{fig:rmse_vs_ws}
\end{figure}

\textcolor{black}{\ref{fig:rmse_vs_ws} presents a comparison of the performances against NDBC buoys for various wind speeds. Unfortunately, collocations are scarce at very low and high wind speeds, which are the intervals where both the GMF and deep learning models have higher errors.}

In the following, we observe two cases where rainfall was detected on the buoy position at the time of observation.

\subsubsection{2017-01-08 01:58:19 at NDBC 46054}

\begin{figure}
    \centering
    \resizebox{\linewidth}{!}{%
    \begin{tabular}{c}
        \includegraphics[width=1.\linewidth]{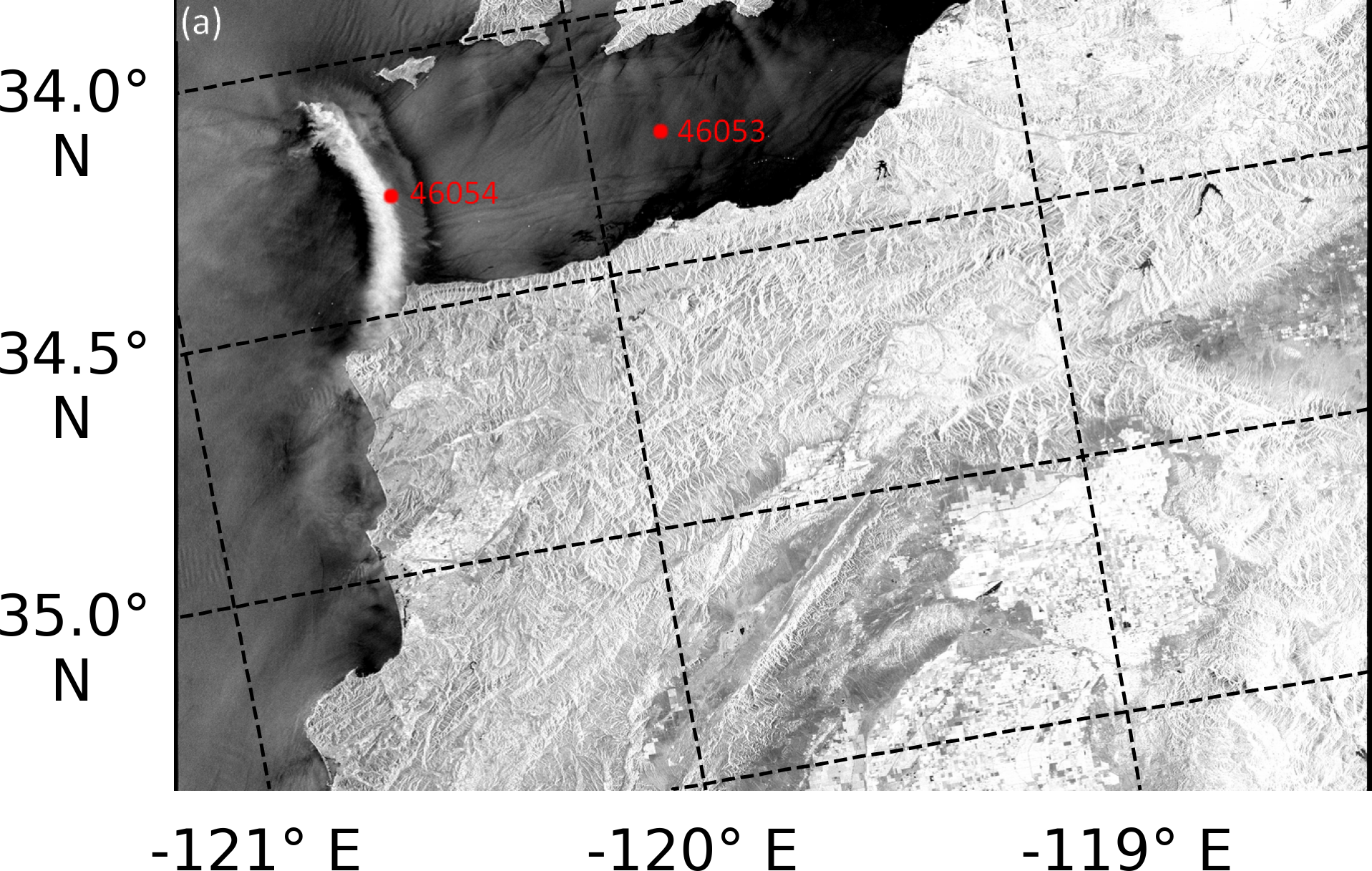} \\
        \includegraphics[width=0.75\linewidth]{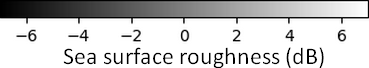}\\
        \includegraphics[width=0.75\linewidth]{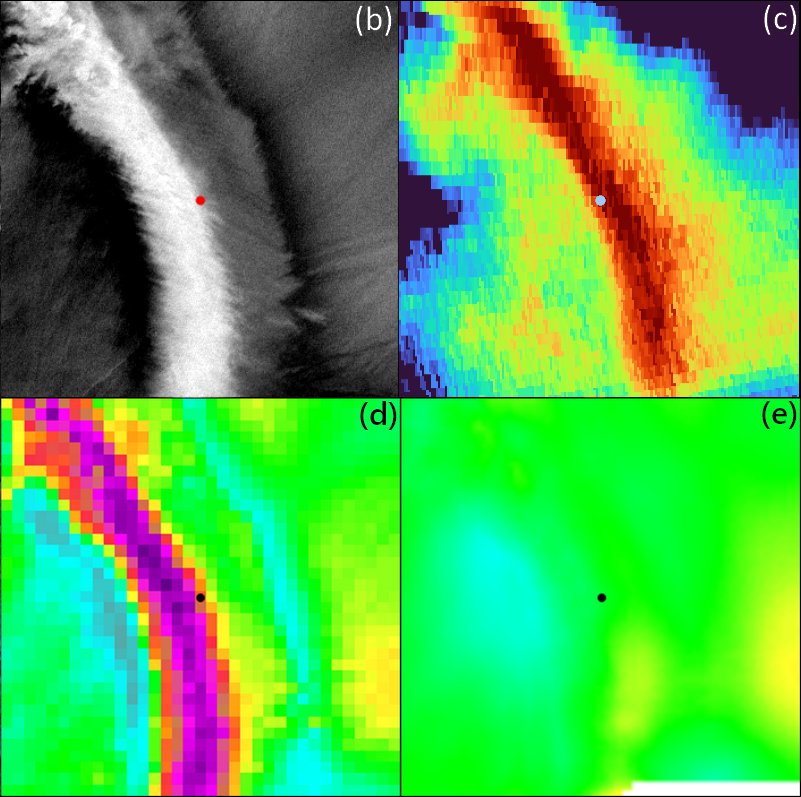} \\
        \includegraphics[width=0.75\linewidth]{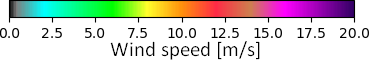}\\
        \includegraphics[width=0.75\linewidth]{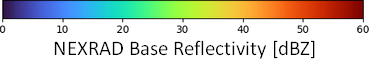}
    \end{tabular}}
    \caption{Sea surface roughness from the January 08\textsuperscript{th} 2017 at 01:58:19 in VV channel (a), zoom on an area of 35x35 km centred on the buoy NDBC 45054 (b), \textcolor{black}{NEXRAD base reflectivity acquired at 01:57:58 (c)}, wind speed given by the GMF (d) and by the deep learning model (e).}
    \label{fig:20170108t015819_square}
\end{figure}

The observation from 2017-01-08 01:58:19 covers the north of the Californian Channel Islands (\ref{fig:20170108t015819_square}.a). Several meteorological buoys are dispersed over the channel, including NDBC 46054 and NDBC 46053, which are indicated as red dots. The wind speed over the area is mostly around 6 \textcolor{black}{m/s}, but a squall line appears at the position of NDBC 46054 and spans over a dozen kilometers. \textcolor{black}{A surge in the air column reflectivity is recorded by weather radar from the NEXRAD network. The ground station is located at 119 km.} The GMF indicates very high wind speeds, higher than 20 m/s (\ref{fig:20170108t015819_square}.d). The deep learning model attenuates these values to between 6 m/s and 8 m/s (\ref{fig:20170108t015819_square}.e).

\begin{figure}
    \centering
    \begin{tabular}{c}
        (a) \\
        \includegraphics[width=0.85\linewidth]{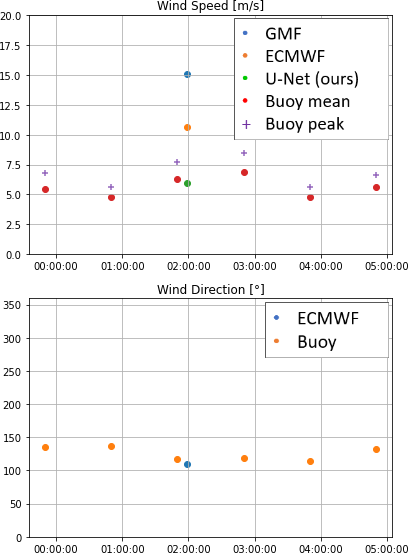} \\
        (b) \\
        \includegraphics[width=0.85\linewidth]{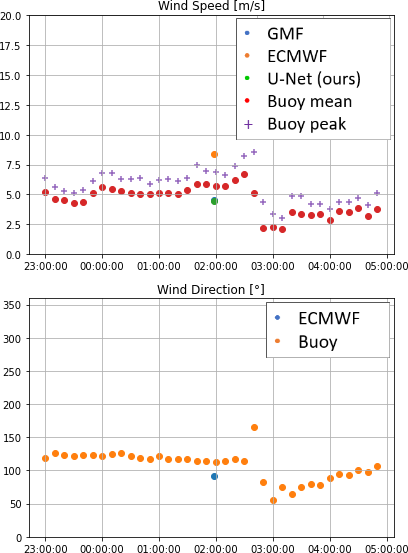}  \\ 
    \end{tabular}
    \caption{Time series of the NDBC buoy wind measurements around January 08\textsuperscript{th} 2017 01:58:19 for NDBC 46054 (a) and NDBC 46053 (b), and the estimation from the GMF, the deep learning model and the atmospheric model.}
    \label{tab:20170108t015819_timeseries}
\end{figure}

For NDBC 46054 (\ref{tab:20170108t015819_timeseries}.a), only one measurement of wind speed and direction per hour is available. It recorded a wind speed of 6.3 m/s eight minutes before the SAR observation. The GMF and the deep learning model estimated wind speeds of 15.1 m/s and 5.9 m/s, respectively. While the temporal resolution of NDBC 46054 is one measurement per hour, NDBC 46053 records data every ten minutes. Furthermore, the gust front appears to be moving toward the right part of the observation. This can be seen in the time series in \ref{fig:20170108t015819_square} as a large variation in wind direction between 02:40:00 and 03:00:00. The variation in wind speed seems to precede the variation in direction, first increasing then decreasing to a lower wind regime. On NDBC 46053 (\ref{tab:20170108t015819_timeseries}.ba), the GMF and the deep learning model agree on a wind speed of 4.5 m/s, which is slightly lower than the in-situ data of 5.4 m/s. Since the distance between NDBC 46054 and NDBC 46053 is approximately 60 km, the progression of the gust front can be estimated to be around 90 km/h. With a width of around 5 or 6 km, the whole system would pass the buoys in three minutes. This means that even NDBC 46053 may not have been able to accurately estimate the wind speed due to its low temporal resolution. However, it is worth noting that even the gust speed at NDBC 46054, defined as the maximum wind speed over a given number of seconds, does not record a speed higher than 9 m/s.

\subsubsection{SAR-20191006T232853 NDBC-41009}

\begin{figure}
    \centering
    \begin{tabular}{c}
        \includegraphics[width=1.0\linewidth]{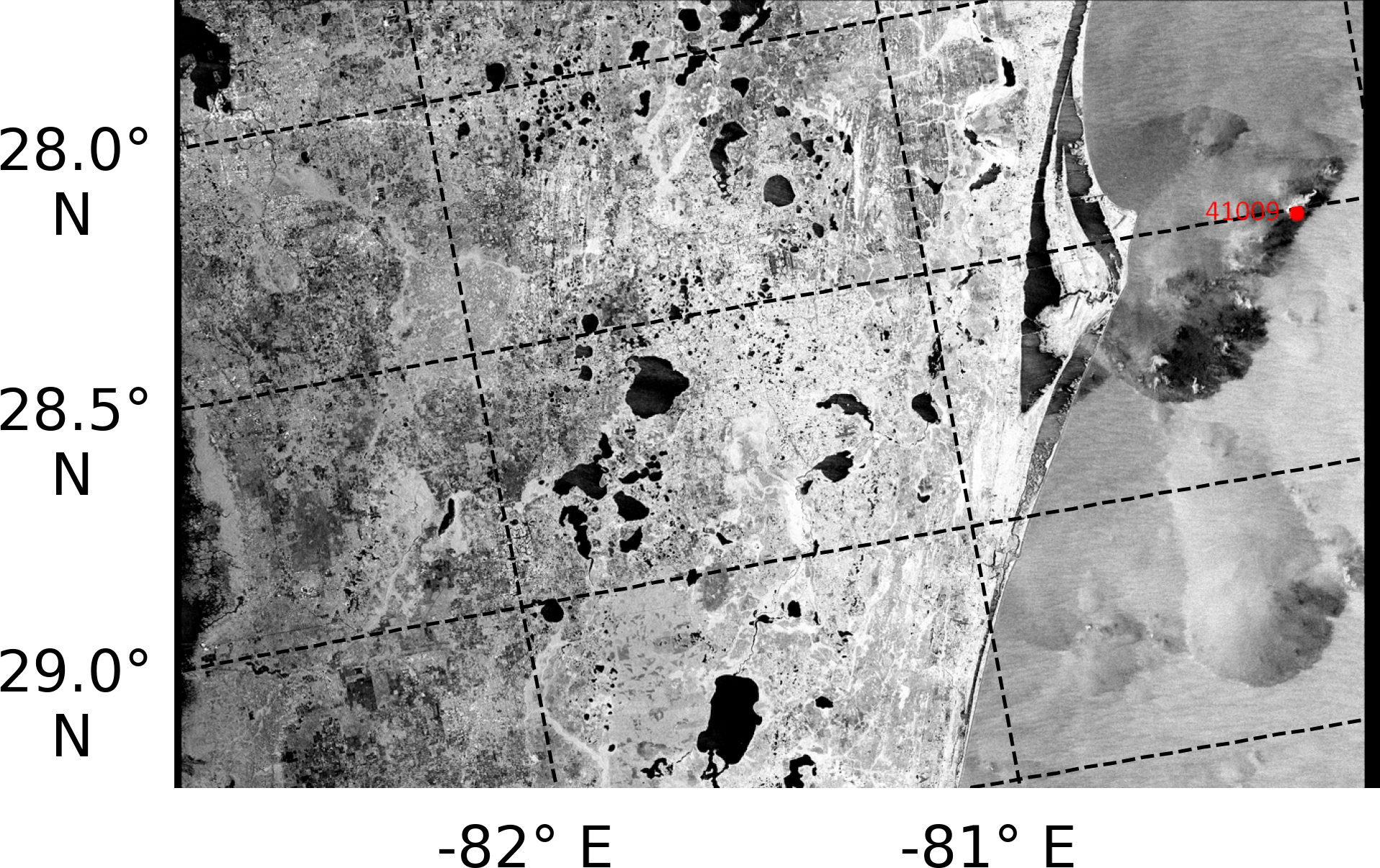}\\
        \includegraphics[width=0.75\linewidth]{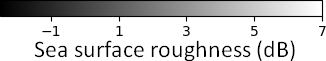}\\
        \includegraphics[width=0.75\linewidth]{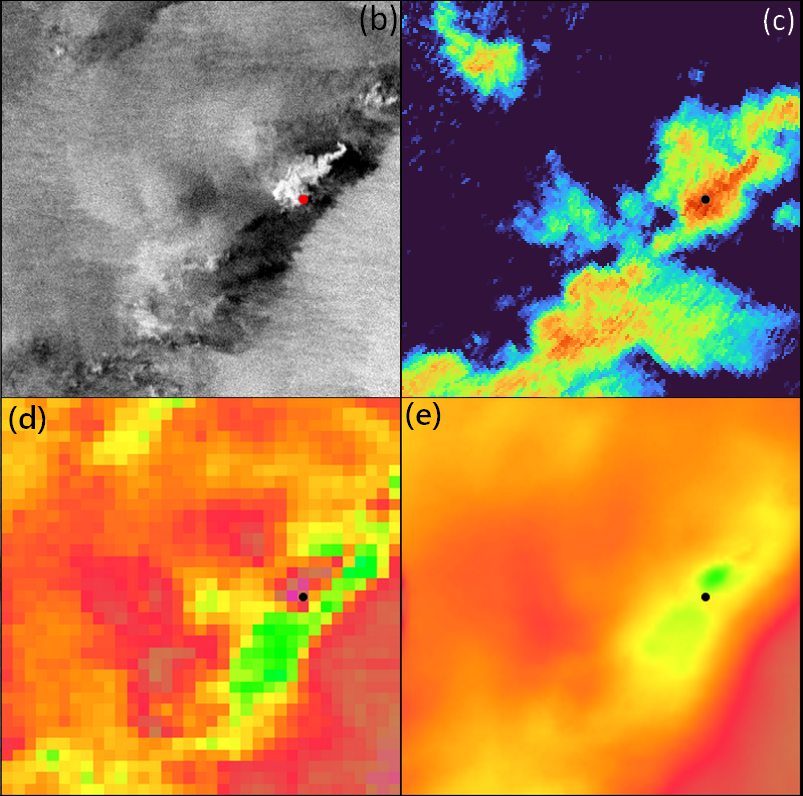}\\
        \includegraphics[width=0.75\linewidth]{src/cmap/ecmwf_cmp.png}\\
        \includegraphics[width=0.75\linewidth]{src/cmap/nexrad_cmap.png}
    \end{tabular}
    \raggedleft
    \caption{Sea surface roughness from October 06\textsuperscript{th} 2019 at 23:28:53 in VV channel in dB (a), zoom on an area of 35x35 km around the buoy NDBC 41009 (b), \textcolor{black}{NEXRAD observation acquired at 23:27:33 (c)}, wind speed given by the GMF (d) and by the deep learning model (e).}
    \label{fig:20191006T232853_41009_square}
\end{figure}

The observation from 2019-10-06 23:28:53 was recorded on the east coast of Florida. While most of the swath covers the marshes around Orlando and Cap Canaveral rather than the ocean, convective precipitation can be observed in the right part of the image (\ref{fig:20191006T232853_41009_square}.a). The cells are moving downward (north-north-east), as indicated by the stronger gradient of the convective front. Since the wind from the convection is opposing the underlying wind regime, an area of lower wind speed appears as an area of lower backscatter. \textcolor{black}{Rainfall was detected by a weather radar from the NEXRAD network located at 64 km (\ref{fig:20191006T232853_41009_square}.c)}. The GMF is impacted by these rain signatures and estimates a very high local wind speed (\ref{fig:20191006T232853_41009_square}.d). The deep learning model is less affected by the rain signatures, but also appears to blur the low wind speed area (\ref{fig:20191006T232853_41009_square}.e).

\begin{figure}
    \centering
    \includegraphics[width=0.85\linewidth]{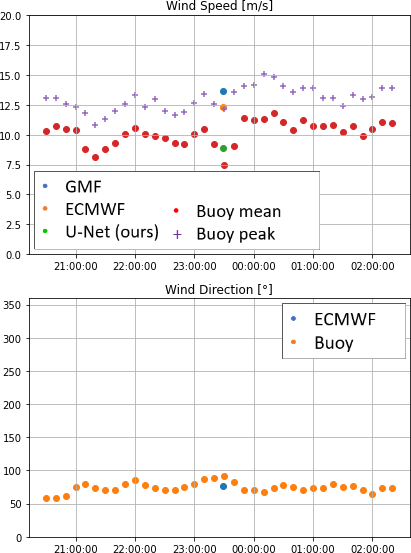}
    \caption{Time series of the NDBC buoy wind measurements around October 06\textsuperscript{th} 2019 at 23:28:53 for NDBC 41009, and the estimation from the GMF, the deep learning model and the atmospherical model.}
    \label{tab:20191006T232853_timeseries}
\end{figure}

The time serie from NDBC 41009 in \ref{tab:20191006T232853_timeseries} indicates that the lower backscattering was indeed caused by a drop in wind speed rather than a change in direction, as the latter does not significantly change during the passage of the convective cell (possibly because the underlying wind regime is strong). It does record a sudden drop in wind speed to 7.5 m/s one minute after the SAR observation, while the GMF and the deep learning model estimated wind speeds of 13.7 m/s and 8.9 m/s, respectively.

\section{Conclusion}

Previous studies have shown that high-resolution rain signatures can be automatically extracted from SAR observations. Using this SAR rainfall segmenter, we built a wind estimation dataset where 50\% of the patches contain rainfall examples. Samples were chosen so that a SAR-based and a SAR-independent wind speed model agree on non-rain pixels, ensuring that their estimates are close to the true wind speed. A UNet architecture was trained on this dataset to estimate wind speeds based on the SAR-independent atmospheric model. We tested several input combinations and found that the most important parameter was the wind speed prior from the geophysical model function, which the deep learning model had difficulty emulating.

Collocations with buoy in-situ measurements show that the model outperforms the current Geophysical Model Function (GMF) on rain areas, reducing the RootMean Square Error (RMSE) by 27\% (resp. 45\%) for rain rates higher than 1 mm/h (resp. 3 mm/h). On rainless areas, performances are similar with a small reduction of the RMSE by 2.7\%. However, since the buoys have a time resolution of ten minutes, some quick sub-mesoscale processes, such as gust fronts, are difficult to register. The limited spatial range of the buoys also makes it challenging to observe rare phenomena. Future work should address these concerns.


\appendix

A secondary dataset was created differing with the main dataset by \ref{eq:dataset_policy}. Here, the balancing policy is defined as:

\begin{equation}
    \forall x, P(x) = P^+(x) = P^-(x)
\end{equation}

\begin{figure}
    \centering
    \includegraphics[width=0.95\linewidth]{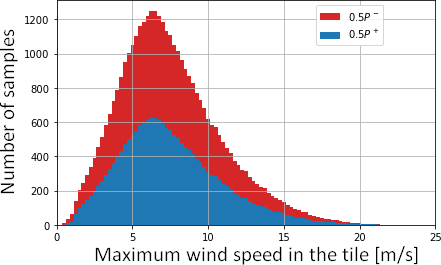}
    \caption{Wind speed distribution for rain (blue) and rainless (red) patches.}
    \label{fig:dataset_wind_distribution_b}
\end{figure}

The distributions $P^+$ and $P^-$ are presented in \ref{fig:dataset_wind_distribution_b}. The balancing is performed for every wind speed, which lead to remove 84\% of the rain patches -especially at 5 and 10 m/s- since the number of rain patches at 8 m/s is limited. 

\textcolor{black}{Models were optimized under the same training process as in subsection \ref{section:Training}, in particular with the same number of weight updates.} Comparison with the first balancing scheme display lower performances despite the more accurate balancing, as indicated \ref{table:vs_buoy_2}.

\begin{table}[!ht]
    \scriptsize
    \centering
    \resizebox{\linewidth}{!}{%
    \begin{tabular}{|cc||c|c|c|}\hline
        && \multirowcell{2}{Dataset\\I} & \multirowcell{2}{Dataset\\II} &GMF \\&&&&\\\hline
        
        \multirowcell{3}{\rotatebox[origin=c]{90}{Bias}}& $< 1 mm/h$ & 
        \textcolor{white}{$\Diamond$} 0.73 [0.04] &  
        \textcolor{white}{$\Diamond$} 0.73 [0.08]  & 
        \textcolor{black}{$\bigstar$} 0.71 
        \\
         & $[1, 3] mm/h$ & 
        \textcolor{black}{$\bigstar$}\bf 1.38 [0.04] &  
        \textcolor{white}{$\Diamond$}\bf 1.40 [0.10]  & 
        \textcolor{white}{$\Diamond$}\bf 2.18  
        \\
        & $> 3 mm/h$ & 
        \textcolor{white}{$\Diamond$}\bf 0.92 [0.07] &
        \textcolor{black}{$\bigstar$}\bf 0.83 [0.14] &  
        \textcolor{white}{$\Diamond$} 2.\textcolor{black}{38}  
        \\\hline
        
        \multirowcell{3}{\rotatebox[origin=c]{90}{RMSE}} & $< 1 mm/h$ & 
        \textcolor{black}{$\bigstar$}\bf 1.44 [0.03] &  
        \textcolor{white}{$\Diamond$}\bf 1.46 [0.06] & 
        \textcolor{white}{$\Diamond$} 1.48  
        \\
         & $[1, 3] mm/h$ & 
        \textcolor{black}{$\bigstar$}\bf 1.81 [0.04] &  
        \textcolor{white}{$\Diamond$}\bf 1.85 [0.10] & 
        \textcolor{white}{$\Diamond$} 2.18  
        \\
        & $> 3 mm/h$ & 
        \textcolor{black}{$\bigstar$}\bf 1.60 [0.10] &  
        \textcolor{white}{$\Diamond$}\bf 1.61 [0.10] & 
        \textcolor{white}{$\Diamond$} 2.93 
        \\\hline
        
        \multirowcell{3}{\rotatebox[origin=c]{90}{PCC}} & $< 1 mm/h$ & 
        \textcolor{black}{$\bigstar$}\bf 93.6\% [0.16\%] &  
        \textcolor{white}{$\Diamond$} 93.3\% [0.15\%]  & 
        \textcolor{white}{$\Diamond$} 93.4\%  
        \\
         & $[1, 3] mm/h$ & 
        \textcolor{black}{$\bigstar$}\bf 96.3\% [0.22\%] &  
        \textcolor{white}{$\Diamond$}\bf 96.1\% [0.23\%]  & 
        \textcolor{white}{$\Diamond$} 95.3\%
        \\
         & $> 3 mm/h$ & 
        \textcolor{black}{$\bigstar$}\bf 95.9\% [0.35\%] &  
        \textcolor{white}{$\Diamond$}\bf 95.0\% [0.46\%] & 
        \textcolor{white}{$\Diamond$} 91.3\%
        \\\hline

    \end{tabular}}
    \caption{Bias, Root Mean Square Error of model \Romannum{4}, the GMF, for each rainfall level. The best result for each rainfall level is indicated by $\bigstar$. Results better than the baseline are in bold. Results of deep learning models are given as mean and standard deviation in brackets.}
    \label{table:vs_buoy_2}
\end{table}


\bibliographystyle{IEEEtran}
\bibliography{main}

\vskip -2\baselineskip plus -1fil

\begin{IEEEbiography}[{\includegraphics[width=1in,height=1in,clip,keepaspectratio]{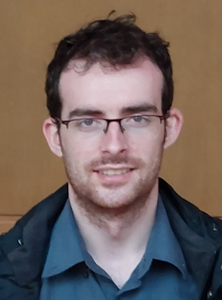}}]{Aurélien COLIN}
After graduating from the IMT Atlantique as an engineer, he began a Ph.D and is currently studying the segmentation of various meteorological and ocean phenomena on Synthetic Aperture Radar using deep learning models.
\end{IEEEbiography}

\vskip -2\baselineskip plus -1fil

\begin{IEEEbiography}[{\includegraphics[width=1in,height=1in,clip,keepaspectratio]{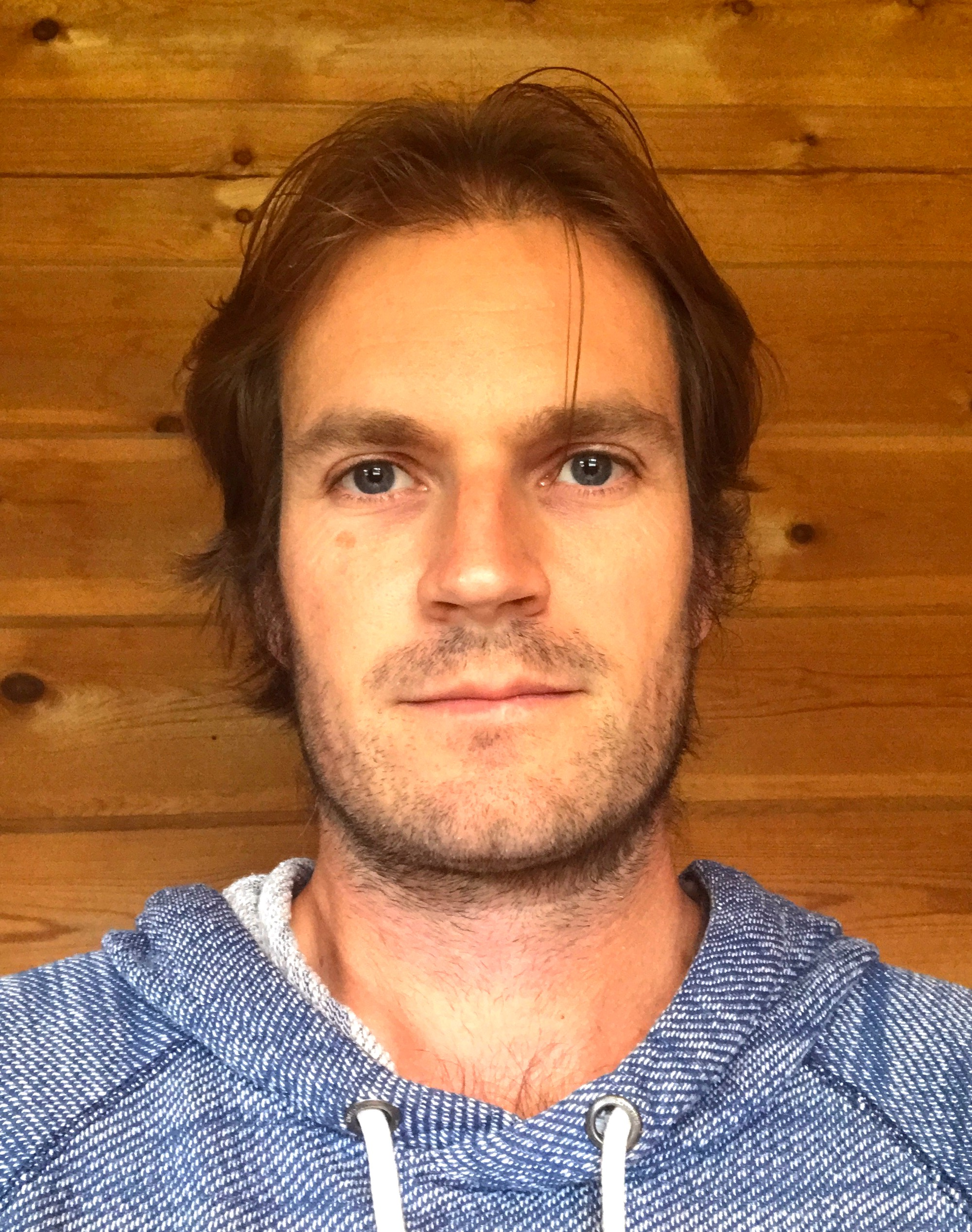}}]{Pierre TANDEO}
was born in France in 1983. He received the M.S. degree in applied statistics from Agrocampus Ouest, Rennes, France, and the Ph.D. degree from the Oceanography from Space Laboratory at IFREMER, Brest, France, in 2010. Then, he spent two years as a Postdoctoral Researcher with the Atmospheric Science Research Group, University of Corrientes, Argentina, and three years at Télécom Bretagne, Brest, France. Since 2015, he is an associate professor at IMT Atlantique, Brest, France, and a researcher at Lab-STICC, CNRS, France. Since 2019, he is an associate researcher at the Data Assimilation Research Team, RIKEN Center for Computational Science, Kobe, Japan. His main research interests are focused on IA, data assimilation, and inverse problems for geophysics.
\end{IEEEbiography}

\vskip -2\baselineskip plus -1fil

\begin{IEEEbiography}[{\includegraphics[width=1in,height=1in,clip,keepaspectratio]{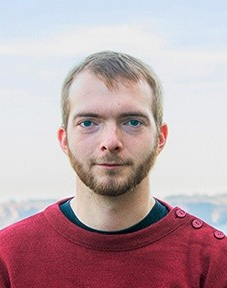}}]{Charles PEUREUX}
is a research engineer at CLS since 2020 where he is mainly involved with SAR oceanography. He previously worked at IFREMER where he defended his thesis on the observation and modelling of the directional properties o short gravity waves. Thereafter, he worked on the SKIM satellite mission, especially for the production of high resolution sea state numeral simulations.
\end{IEEEbiography}

\vskip -2\baselineskip plus -1fil

\begin{IEEEbiography}[{\includegraphics[width=1in,height=1in,clip,keepaspectratio]{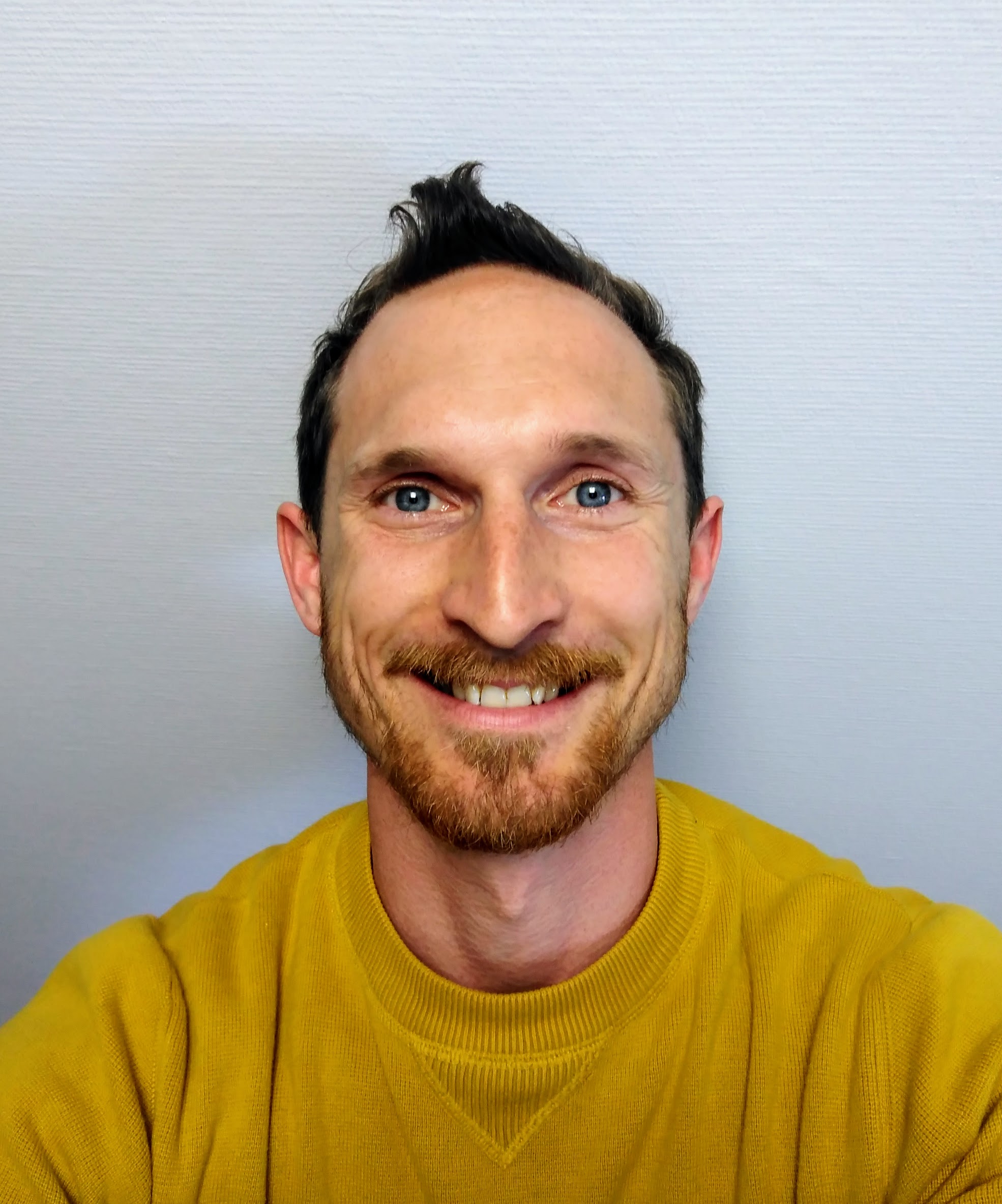}}]{Romain HUSSON}
is a scientist and project engineer at CLS, in the Environmental Monitoring and Climate Business Unit. After working as intern at NASA JPL and as YGT at ESRIN on SAR wind-wave-current activities, he completed in 2012 his PhD in Physical Oceanography at CLS Brest developing methods to estimate synthetic swell field from SAR wave mode observations. Since then, he has been involved in several applicative and R\&D projects such as wind/wave products for marine renewable energy and contributed to several ESA and European projects such as Cal/Val activities for ENVISAT and S-1 mission performance center as Level-2 expert. 
\end{IEEEbiography}

\vskip -2\baselineskip plus -1fil

\begin{IEEEbiography}[{\includegraphics[width=1in,height=1in,clip,keepaspectratio]{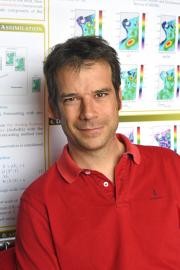}}]{Ronan FABLET}
is a Professor at IMT Atlantique and a research scientist at Lab-STICC in the field of Data Science and Computational Imaging. He is quite involved in interdisciplinary research at the interface between data science and ocean science, especially space oceanography and marine ecology. His current research interests include deep learning for dynamical systems and applications to the understanding, analysis, simulation and reconstruction of ocean dynamics, especially using satellite ocean remote sensing data.
\end{IEEEbiography}

\end{document}